\documentclass{article} 
\usepackage{iclr2026_conference,times}

\usepackage{amsmath,amsfonts,bm}

\def\eqref#1{equation~\ref{#1}}

\def\1{\bm{1}}

\DeclareMathAlphabet{\mathsfit}{\encodingdefault}{\sfdefault}{m}{sl}
\SetMathAlphabet{\mathsfit}{bold}{\encodingdefault}{\sfdefault}{bx}{n}

\usepackage{hyperref}
\usepackage{url}

\usepackage{subfigure}
\usepackage{graphicx}
\usepackage{amssymb}
\usepackage{tabularx}
\usepackage{wrapfig}
\usepackage{threeparttable}
\usepackage{booktabs}
\usepackage{multirow}
\usepackage[normalem]{ulem}
\useunder{\uline}{\ul}{}
\usepackage{tcolorbox}
\tcbuselibrary{breakable}
\usepackage[T1]{fontenc}    

\usepackage[ruled,linesnumbered]{algorithm2e}

\title{Reasoning Scaffolding: Distilling the Flow of Thought from LLMs}

\iclrfinalcopy

\author{
\textbf{Xiangyu Wen\textsuperscript{1}\thanks{Email: xywen22@cse.cuhk.edu.hk}},
\textbf{Junhua Huang\textsuperscript{2}},
\textbf{Zeju Li\textsuperscript{1}},
\textbf{Min Li\textsuperscript{3}},
\textbf{Jianyuan Zhong\textsuperscript{1}},
\textbf{Zhijian Xu\textsuperscript{1}},\\
\textbf{ Mingxuan Yuan\textsuperscript{2}},
\textbf{Yongxiang Huang\textsuperscript{4}},
\textbf{Qiang Xu\textsuperscript{1}\thanks{Corresponding author.}},\\
\\
 \textsuperscript{1}The Chinese University of Hong Kong\quad
 \textsuperscript{2}HUAWEI Noah's Ark Lab\quad
 \textsuperscript{3}Southeast University\\
 \textsuperscript{4}HUAWEI Hong Kong Research Center
}

\begin{document}

\maketitle

\begin{abstract}

The prevailing approach to distilling reasoning from Large Language Models (LLMs)—behavioral cloning from textual rationales—is fundamentally limited. It teaches Small Language Models (SLMs) to mimic surface-level patterns rather than the underlying algorithmic structure of thought, resulting in a critical lack of logical robustness. We argue that instead of cloning text, distillation should transfer this algorithmic structure directly. We introduce \textbf{Reasoning Scaffolding}, a framework that reframes reasoning as a structured generation process. Our method first abstracts the teacher's thought process into a sequence of discrete, interpretable semantic signals (e.g., \texttt{Contrast}, \texttt{Addition}) that act as a scaffold. The student model is then trained via a multi-task objective to both (1)~\emph{predict} the next semantic signal, anticipating the reasoning flow, and (2)~\emph{generate} the corresponding step, conditioned on that signal. This multi-task scheme acts as a powerful regularizer, compelling the student to internalize the computational patterns of coherent reasoning. On a suite of challenging reasoning benchmarks, our method significantly outperforms state-of-the-art distillation in both accuracy and logical consistency, providing a path towards creating smaller models that are genuine reasoners, not just fluent mimics\footnote{Our code is available at: \href{https://anonymous.4open.science/r/ReasoningScaffolding-7F1F}{https://anonymous.4open.science/r/ReasoningScaffolding-7F1F}}.


\end{abstract}

\section{Introduction}
\label{sec:introduction}

The prevailing approach to distilling reasoning from Large Language Models (LLMs)---behavioral cloning from Chain-of-Thought (CoT) rationales~\citep{wei2022chain}---is fundamentally limited \citep{li2022explanations, Ho2022LargeLM, shridhar2023distilling}. This method treats reasoning as a text imitation task, effectively forcing Small Language Models (SLMs) into a form of rote memorization \citep{gu2024minillm, anil2023palm}. While this can teach stylistic fluency, it fails to transfer the underlying \emph{algorithmic structure} of the teacher's thought process. Consequently, the resulting student models are often brittle, producing arguments that are logically inconsistent or nonsensical when faced with novel problems \citep{shridhar2023distilling}.

To move beyond this superficial mimicry, we argue for a paradigm shift: instead of teaching a model what to write, we must teach it how to think. Our key insight is that the teacher's reasoning process can be abstracted \emph{from verbose text into its core structural blueprint}. This blueprint, composed of discrete, interpretable semantic signals like \texttt{Contrast} or \texttt{Elaboration}, governs the flow of a coherent argument. We introduce \textbf{Reasoning Scaffolding}, a new pedagogical framework that distills this structural blueprint, providing the student model with a scaffold to construct its own robust reasoning.

Our framework implements this principle through a novel multi-task training regimen. We teach the SLM to reason step-by-step by having it simultaneously learn two crucial skills: (1) to \emph{anticipate} the flow of a logical argument by predicting the next semantic signal, and (2) to \emph{execute} a specific reasoning move by generating the corresponding text, conditioned on that signal. This dual objective, which uses the signal prediction task as a powerful regularizer for logical coherence, compels the student to internalize the computational patterns of reasoning rather than simply cloning text.

Our contributions are as follows:
\begin{itemize}
\item We introduce Reasoning Scaffolding, a new pedagogical framework that distills the structured, algorithmic flow of a teacher's reasoning, moving beyond surface-level text imitation.
\item We propose a principled method for extracting and categorizing a `reasoning scaffold' from textual rationales, creating a structured and interpretable training signal.
\item We demonstrate through extensive experiments on benchmarks like GSM8K \citep{cobbe2021gsm8k} and StrategyQA \citep{Geva2021DidAU} that our method yields SLMs that are significantly more accurate and logically robust than those trained with state-of-the-art distillation methods.
\item We provide in-depth analysis showing closer alignment between the student and teacher's logical representations, providing strong evidence that our method produces genuine reasoners, not just fluent mimics.
\end{itemize}

In summary, this work establishes a theoretically grounded and empirically validated framework for transferring the core reasoning ability of LLMs, advancing the goal of creating powerful, efficient, and truly capable SLMs.
\section{Related Work}
This section reviews the key research areas that underpin our work: (1) knowledge distillation in language models, (2) approaches for distilling reasoning capabilities, and (3) alternative paradigms for incorporating structure and logic into text generation.

\subsection{Knowledge Distillation from LLMs}

Knowledge Distillation (KD) \citep{hinton2015distilling, fang2025knowledge, xu2024surveyknowledgedistillationlarge} is a foundational technique for making the capabilities of large language models (LLMs) accessible and practical. The central goal is to transfer knowledge from a powerful, resource-intensive teacher model to a smaller, more efficient student model, thereby enabling wider deployment and application. Recent surveys \citep{xu2024surveyknowledgedistillationlarge, zhu2024surveycompression} show that KD in the LLM era now covers a broad spectrum of approaches, including model compression, skill transfer, and self-improvement for open-source models.

Recent advances have adapted core distillation algorithms to better suit the generative nature of LLMs. Traditional KD methods \citep{wang2025abkd} often use forward Kullback-Leibler (KL) divergence, which can cause the student to overestimate rare outputs from the teacher. To address this, \citet{gu2024minillm} propose using reverse KL divergence, which penalizes the student for generating outputs the teacher considers unlikely—proving more effective for generative tasks. Another major challenge is the architectural and vocabulary mismatch between teacher and student models. The Dual-Space Knowledge Distillation (DSKD) framework \citep{zhang2024dualspaceknowledgedistillationlarge} tackles this by aligning representations through a unified output space and cross-model attention, enabling distillation even when models use different tokenizers and vocabularies.

Beyond algorithmic improvements, recent work has focused on enhancing both the distillation process and the training data. For example, TinyLLM \citep{dai2024beyond} introduces multi-teacher distillation, allowing students to learn from a diverse set of teacher LLMs and acquire a richer set of skills and reasoning patterns. Other approaches emphasize distilling not just final answers, but also intermediate rationales or Chain-of-Thought steps, to impart deeper reasoning abilities \citep{ma2025slowtuninglowentropymasking}. Research also shows that generating training data more efficiently—such as by incorporating student feedback into the teacher’s output—can greatly reduce the need for large amounts of synthetic data. Additionally, curriculum-based fine-tuning schedules have been proposed to help SLMs internalize symbolic knowledge and perform complex reasoning without external tools at inference time \citep{liao2024textitskinterninternalizingsymbolicknowledge}.

However, despite these advancements, most existing KD methods focus on behavioral cloning from textual rationales. They prioritize transferring knowledge and stylistic fluency, while often overlooking the deeper reasoning capabilities and algorithmic paradigms that underpin robust logical thinking.

\subsection{Reasoning Capability Distillation from LLMs}

In addition to knowledge distillation, recent research seeks to transfer the advanced reasoning abilities of Large Language Models (LLMs)—especially those elicited by Chain-of-Thought (CoT) prompting~\citep{wei2022chain,wang2023selfconsistency,Wang2023PlanandSolvePI}—to smaller, more efficient models (SLMs) \citep{srivastava2025reasoningabilitysmalllanguage}. The standard approach, reasoning distillation, involves fine-tuning SLMs on multi-step rationales generated by LLMs \citep{shridhar2023distilling, Ho2022LargeLM, wang2025efficientlongcotreasoning}. However, simply mimicking teacher rationales often leads SLMs to learn superficial patterns rather than the underlying logic, resulting in limited robustness.

To address this, recent works have focused on improving distillation data and methodology. For example, EDIT \citep{Ho2022LargeLM} uses pairs of similar but outcome-divergent reasoning traces to highlight critical inference steps, while other studies prune redundant steps from CoTs to promote concise, effective reasoning \citep{wang2025efficientlongcotreasoning, li2025compressingchainofthoughtllmsstep}. Mentor models \citep{lee2024mentor}, rationale decomposition \citep{xi2024learning}, and modular architectures \citep{hinton2015distilling} further enhance transfer by providing structured or higher-quality supervision.

However, most reasoning distillation methods overlook the underlying structural patterns present in extended reasoning traces—patterns that provide valuable signals for interpreting and guiding LLM reasoning. In the realm of interpretability, Concept Bottleneck LLMs (CB-LLMs) \citep{sun2025concept} introduce a framework where token decoding is made transparent through a concept bottleneck layer, allowing users to trace specific task concepts, such as `Technology' and `Business', while maintaining competitive accuracy. Yet, CB-LLMs primarily focus on token-level decoding in classification and plain text generation tasks. Building on this foundation, we extend the concept bottleneck approach by introducing discrete semantic reasoning signals into step-by-step, challenging reasoning tasks. This enables the transfer of algorithmic reasoning structure directly, rather than merely cloning textual rationales, addressing both interpretability and logical robustness in distillation.
\section{Method}

In this section, we detail the design and implementation of the Reasoning Scaffolding framework. Our approach comprises three key components: (1) \textbf{Logic Representation Distillation}, which prepares the training data by abstracting the teacher's reasoning process into structured semantic signals; (2) \textbf{Joint Training of Reasoning Proposer and Semantic Signal Predictor}, which enables the student model to anticipate and generate each step of reasoning; and (3) \textbf{Semantic Signal-Guided Reasoning at Inference}, which leverages the learned scaffold to guide the model's reasoning process during inference.

\subsection{Logic Representation Distillation}
\label{sec:data}

\vspace{-5pt}
\begin{figure*}[ht]
    \centering
    \makebox[\textwidth][c]{\includegraphics[width=0.85\textwidth]{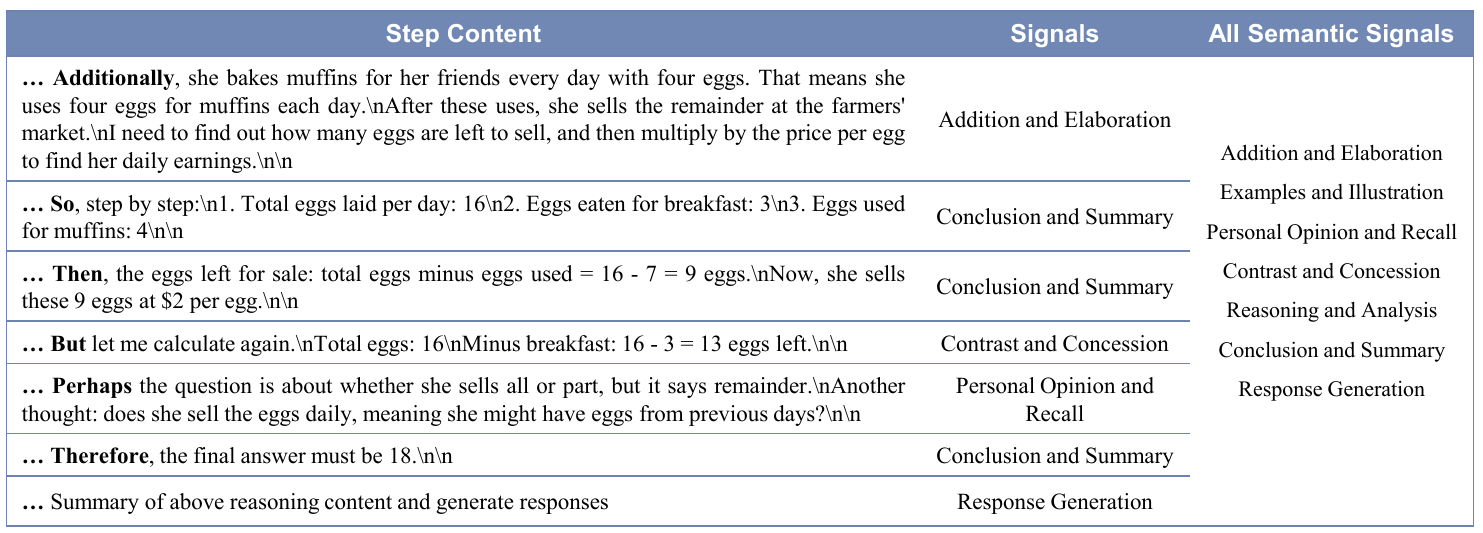}}
    \caption{Examples from GSM8K illustrating clear logical transition signals that guide reasoning direction.}
    \label{fig:motivation}
\end{figure*}
\vspace{-5pt}

We first query a large reasoning model (LRM, e.g., Deepseek-R1) using zero-shot prompting to obtain detailed reasoning traces (the prompt can be found in Appendix~\ref{App:promptDG}). This yields a collection of long-thinking examples $S$, each comprising a step-by-step trace and a final summary. As shown in Figure~\ref{fig:motivation}, we can observe that certain keywords—such as `wait', `but', `ok', and `in addition'—naturally signal transitions in reasoning. For example, `in addition' typically introduces supplementary information or elaboration. To systematically capture this phenomenon, we group these keywords into 7 semantic categories, such as `Contrast and Concession', as semantic signals. We selected these 7 signals to ensure both \textbf{internal coherence}, where keywords within a group are semantically related (like `in conclusion' and `therefore'), and \textbf{external comprehensiveness}, where the categories collectively encompass the vast majority of transitions, as confirmed by manual review. A complete list of signals is provided in Table~\ref{tab:signals} in Appendix~\ref{tab:signals}. 

Based on the above initial processing on the reasoning trace and semantic signals. We can build up our Reasoning Scaffolding dataset. We begin by segmenting the initial long-thinking trace into individual steps ($S_i = [A_1, \dots, A_N]$), using double newline characters (`\textbackslash n\textbackslash n') as delimiters. As for generating the corresponding semantic signals for reasoning steps, we adopt a two-stage approach: First, we assign initial signal labels to each individual step based on keyword matches using Table~\ref{tab:signals}. Second, we use a strong LLM (e.g., GPT-4.1) for semantic validation—verifying consistency between each step and its signal (the prompt example is in Appendix~\ref{App:promptSignal}). If a mismatch is detected, the LLM assigns the correct signal. For steps starting without signal keywords, the LLM directly provides the semantic label. The interplay and effectiveness of keyword matching versus LLM-based semantic labeling are further analyzed in Section~\ref{sec:exp:signalPredictor} and Appendix~\ref{app:signalLabeling}.

Using the step-by-step reasoning traces and the defined semantic signals, we generate two sets of training data: (1) pairs for the signal predictor, $\{Q+[A_1, \dots, A_t], \text{Signal}\}$, for predicting next semantic signal, (2) pairs for the proposer model, $\{Q+[A_1, \dots, A_t], \text{Signal}+A_{t+1}\}$, for generating next reasoning step. This process yields robust datasets for training both the signal predictor and the proposer model.

\subsection{Training SLMs as Step-by-Step Proposer and Logic Signal Predictor}
\label{sec:training}

\begin{figure*}[ht]
    \centering
    \makebox[\textwidth][c]{\includegraphics[width=0.85\textwidth]{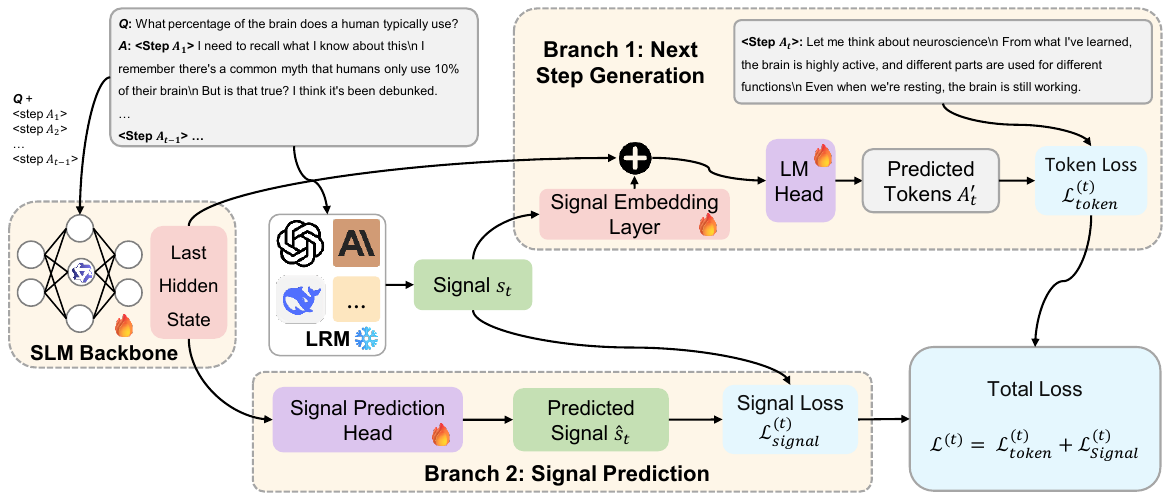}}
    \caption{Reasoning Scaffolding framework. The dual-branch architecture trains an SLM to learn reasoning logic via two simultaneous tasks: signal-guided generation and signal prediction. The model is optimized to learn both the content and the structure of reasoning.}
    \label{fig:pipeline}
\end{figure*}

With the labeled data, we train the SLM to perform signal-guided, step-by-step reasoning. As shown in Figure~\ref{fig:pipeline}, the modified model structure consists of a backbone and two branches. 

The backbone is responsible for processing the input context, and producing the vectors of hidden states. In branch 1, we add an additional signal embedding layer (SEL) before the original language modeling (LM) head. The SEL layer is used to encode the pre-generated semantic signals $S_{i+1}$ into embeddings. The signal embeddings are fused with the backbone's last hidden state through simple addition and then passed to the LM head for next-token prediction. All tokens in a step share the same semantic signal, and training uses a modified next-token prediction loss:

\vspace{-10pt}
\begin{equation}
\mathcal{L}^{(t)} = -\frac{1}{N_t} \sum_{i=1}^{N_t} \log P_{\theta}\left(A_{t,i} \mid A_{<t},\ A_{t,<i},\ s_t\right)
\label{eq:tokenprediction}
\end{equation}
where $A_t$ represents tokens at step $t$ with length of $N_t$, $A_{<t}$ are previous steps, and $s_t$ is the associated semantic signal.

To further align the model’s internal reasoning with the signal structure, we introduce a Signal Prediction Head as Branch 2. This branch compels the backbone to predict the current semantic signal, thereby increasing the model’s sensitivity to signal cues and improving the semantic consistency between each reasoning step and its guiding signal. The signal prediction loss is defined as:

\vspace{-10pt}
\begin{equation}
\mathcal{L}_{\text{signal}}^{(t)} = - \sum_{j=1}^{C} s_{t,j} \log P_{\theta}\left(\hat{s}_{t,j} \mid A_{<t}\right)
\label{eq:signalloss}
\end{equation}
where $C$ is the number of semantic signals. The overall training objective is $\mathcal{L}^{(t)} = \mathcal{L}_{\text{token}}^{(t)} + \mathcal{L}_{\text{signal}}^{(t)}$.

In real-world applications, the signal predictor must generate next semantic signals in place of the ground-truth labels. The updated backbone provides a strong initialization for this predictor. For cold-start scenarios, we initialize a separate SLM with the trained backbone and then focus its training on signal prediction, further enhancing its ability to anticipate semantic signals during inference.

\subsection{Semantic Signal-Guided Reasoning}
\label{sec:inference}

During inference, reasoning at each step is guided by predicted semantic signals. Given a question and the current reasoning trace, the signal predictor infers the most probable semantic signal for the next step. To ensure the correctness and reliability of these signals, we adopt an adaptive strategy. This approach begins by computing the confidence of the predicted signal:

\vspace{-10pt}
\begin{equation}
\text{conf} = \exp\left( \frac{1}{L_t} \sum_{l=1}^{L_t} \log P_{\phi}(s_{t,l} \mid A_{<t}, s_{t,<l}) \right)
\label{eq:signal_logprob}
\end{equation}
where $L_t$ is the length of signal $s_t$ at step $t$.

Predicted signals with confidence exceeding a threshold $\tau$ are used to guide the next reasoning step. If the confidence falls below $\tau$, the predicted signals are considered unreliable for further guidance. In such cases, we terminate the reasoning trace and prompt the model to generate a conclusion using a special `Response Generation' signal. The impact of the signal predictor's sensitivity to the hyper-parameter $\tau$ is analyzed and illustrated in Table~\ref{fig:signalTau}.

\vspace{-10pt}
\begin{figure*}[ht]
    \centering
    \makebox[\textwidth][c]{\includegraphics[width=0.7\textwidth]{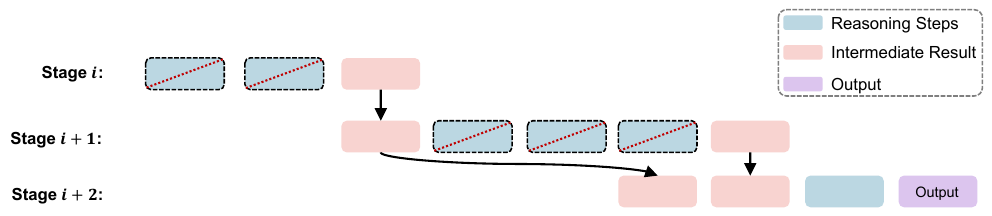}}
    \caption{Token efficiency: By pruning reasoning steps before those labeled `Conclusion and Summary', only intermediate results are retained.}
    \label{fig:reduceToken}
\end{figure*}

Thanks to the interpretability of semantic signals, we can further optimize the reasoning trace. As illustrated in Figure~\ref{fig:reduceToken}, steps labeled `Conclusion and Summary' denote key intermediate results. We prune all other reasoning steps within the same stage, retaining only these essential outputs. This process is repeated iteratively until the final answer is reached, significantly reducing token usage while preserving the critical information needed for downstream reasoning.
\section{Experiments}

\subsection{Experimental Settings}
We evaluate our models on several distinct QA and mathematical benchmarks, each designed to target specific logical reasoning skills. \textbf{StrategyQA}~\citep{Geva2021DidAU} tests implicit multi-step reasoning, requiring the inference of necessary intermediate steps. \textbf{CommonsenseQA}~\citep{talmor-etal-2019-commonsenseqa} assesses the application of common-sense knowledge in a multiple-choice format. \textbf{TruthfulQA}~\citep{Lin2021TruthfulQAMH} measures a model's ability to avoid common misconceptions and imitative falsehoods. \textbf{GSM8K}~\citep{cobbe2021gsm8k} evaluates proficiency in grade-school mathematical problem solving, focusing on multi-step arithmetic and reasoning required to arrive at correct answers. Finally, \textbf{MATH}~\citep{hendrycksmath2021,lightman2023let} assesses advanced mathematical reasoning and problem-solving skills across a broad range of topics, requiring models to generate detailed solutions to challenging competition-level math questions.

For evaluation, we use the Pass@1 metric to compare performance across different experimental settings throughout our paper. We experiment with a series of base models: Qwen-2.5-\{0.5B, 7B, 14B\}-instruct~\citep{Yang2024Qwen25TR}. All experiments are conducted using NVIDIA L40 GPUs and Intel(R) Xeon(R) Gold 6426Y CPUs.

\vspace{-5pt}
\subsection{Main Results}
\vspace{-5pt}

We compare our method against several baselines implemented on different sizes of Qwen models. Specifically, the baselines include: the original pre-trained model, SFT models fine-tuned with Chain-of-Thought (CoT) and Long Thinking data, as well as Long-Thinking models distilled from Deepseek-R1. 

\vspace{-10pt}
\begin{table*}[ht]
\centering
\begin{threeparttable}
\scriptsize
\renewcommand{\arraystretch}{1.1}
\caption{Main results}
\begin{tabular}{ccccccc}
\hline
\textbf{Methods}                       & \textbf{Models}  & \textbf{StrategyQA} & \textbf{CommonsenseQA} & \textbf{TruthfulQA} & \textbf{GSM8K} & \textbf{MATH-500} \\ \hline
\multirow{3}{*}{Original}              & Qwen2.5-0.5B-Ins & 0.543               & 0.475                  & 0.268               & 0.379          & 0.335             \\
                                       & Qwen2.5-7B-Ins   & 0.726               & 0.785                  & 0.706               & 0.875          & 0.738             \\
                                       & Qwen2.5-14B-Ins  & 0.755               & 0.785                  & 0.750               & 0.921          & 0.764             \\
\multirow{3}{*}{CoT SFT}               & Qwen2.5-0.5B-Ins & 0.569               & 0.457                  & 0.682               & 0.394          & 0.351             \\
                                       & Qwen2.5-7B-Ins   & 0.751               & 0.808                  & 0.752               & 0.887          & 0.851             \\
                                       & Qwen2.5-14B-Ins  & 0.760               & 0.810                  & 0.831               & 0.928          & 0.882             \\
\multirow{3}{*}{Long Thinking SFT\tnote{a}}     & Qwen2.5-0.5B-Ins & 0.571               & 0.463                  & 0.670               & 0.412          & 0.388             \\
                                       & Qwen2.5-7B-Ins   & 0.759               & 0.817                  & 0.771               & 0.862          & 0.879             \\
                                       & Qwen2.5-14B-Ins  & 0.768               & 0.845                  & 0.812               & 0.931          & 0.901             \\
\multirow{2}{*}{Long Thinking Distill\tnote{b}} & Qwen2.5-7B-Ins   & 0.680               & 0.670                  & 0.544               & 0.867          & 0.883             \\
                                       & Qwen2.5-14B-Ins  & 0.811               & 0.805                  & 0.763               & 0.936          & 0.904             \\ \hline
\multirow{3}{*}{\textbf{Ours}}         & Qwen2.5-0.5B-Ins & 0.659               & 0.601                  & 0.861               & 0.488          & 0.417             \\
                                       & Qwen2.5-7B-Ins   & 0.832               & 0.866                  & 0.879               & 0.899          & 0.922             \\
                                       & Qwen2.5-14B-Ins  & 0.858               & 0.887                  & 0.917               & 0.942          & 0.928             \\ \hline
\end{tabular}
\begin{tablenotes}
  \scriptsize
  \item[a] Official distilled model released by Deepseek.
  \item[b] Our SFT model trained on long-thinking data distilled and generated by Deepseek-R1.
\end{tablenotes}
\label{tab:mainresults}
\end{threeparttable}
\end{table*}

The experimental results in Table~\ref{tab:mainresults} demonstrate the effectiveness of our method across a variety of question-answering and mathematical reasoning benchmarks. Compared to the original base models, our approach achieves a substantial improvement, with an average increase of approximately 14\% in Pass@1 accuracy across all tasks. Furthermore, when compared to models fine-tuned with Chain-of-Thought (CoT) or Long-Thinking data, our method yields an average improvement of about 8\%.

We also observe a consistent trend of performance improvement as the model size increases. Our method is effective across all model scales, from the smallest 0.5B model to the largest 14B model, highlighting its strong generalization ability to different model capacities and diverse benchmarks.

Notably, our approach brings significant gains to smaller models. For instance, on the TruthfulQA dataset, the 0.5B model's Pass@1 accuracy increases dramatically from around 27\% (original) and 68\% (CoT SFT) to over 86\% with our method. This demonstrates the potential of our approach to substantially enhance reasoning and factual correctness even in resource-constrained settings.

\vspace{-5pt}
\subsection{Comparison of Reasoning Traces}
\vspace{-5pt}

To explicitly demonstrate the advantages of our Reasoning Scaffolding method over the baselines, we compare the reasoning traces generated by different models, including the original model, the CoT-SFT model, and our model enhanced with Reasoning Scaffolding.

\begin{figure*}[ht]
    \centering
    \makebox[\textwidth][c]{\includegraphics[width=1.0\textwidth]{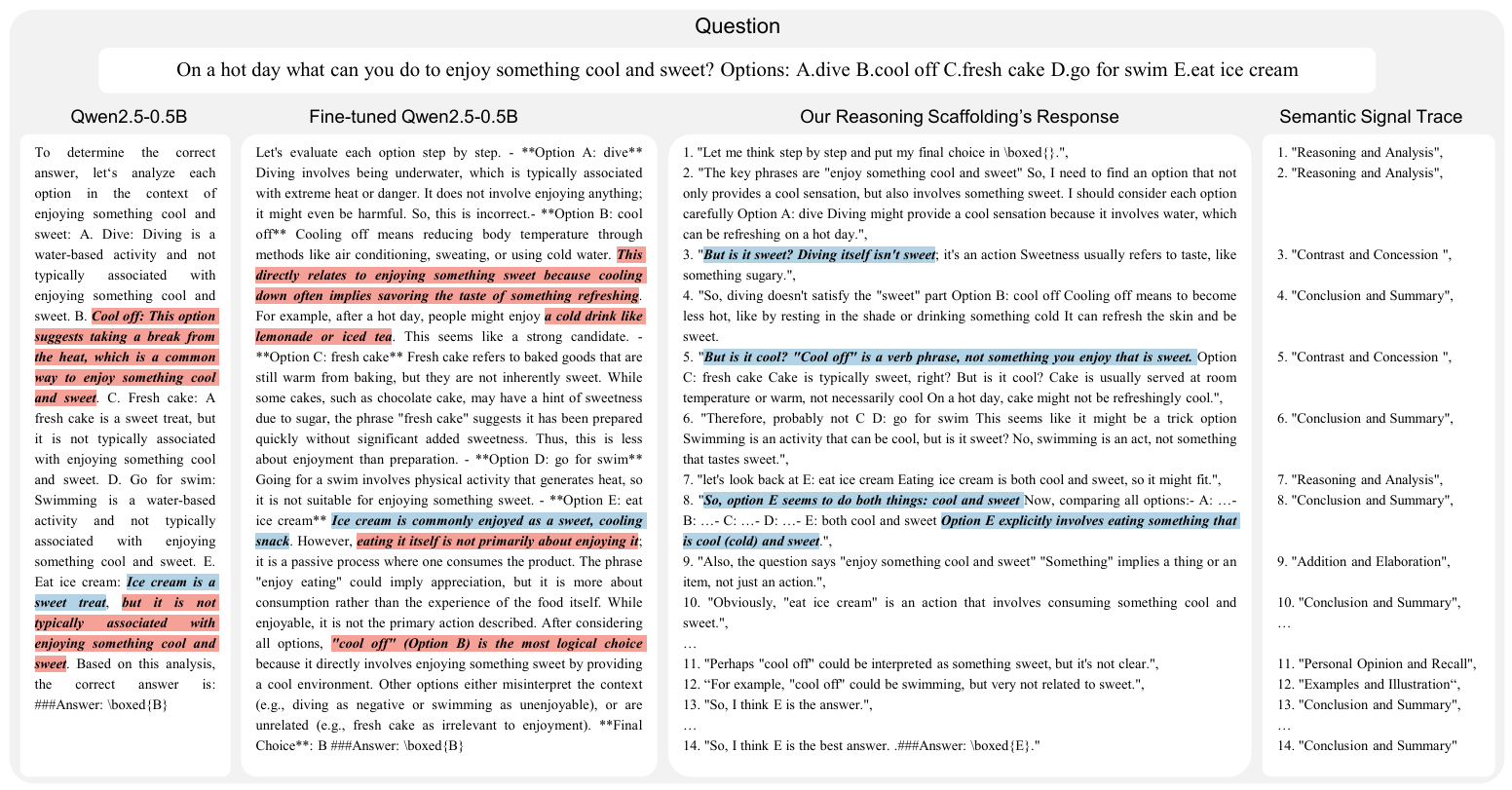}}
    \caption{Case studies analyzing how semantic signal-guided reasoning enhances the model's reasoning capability.}
    \label{fig:caseanalysis}
\end{figure*}

Figure~\ref{fig:caseanalysis} presents detailed reasoning traces produced by different models. These cases illustrate both why and how our approach improves the model's reasoning abilities. 

On the left, the reasoning trace generated by the original Qwen2.5-0.5B model contains clear mistakes. For example, the model incorrectly treats `cool off' as an object and selects it as a possible answer. Although it correctly identifies that `ice cream' is sweet, it fails to establish the connection between `ice cream' and `enjoy something,' resulting in an incorrect final answer.

The CoT-SFT model, fine-tuned on Chain-of-Thought data, also fails to predict the correct answer and continues to consider `cool off' as an option. While it demonstrates some improvement by recognizing that `ice cream' is both sweet and cool, it still does not link eating `ice cream' with the concept of `enjoying something.'

In contrast, our Reasoning Scaffolding method enables the model to correctly identify that `cool off' is not an object, and successfully explains why `ice cream' is the correct answer. These results highlight how our approach strengthens the model's ability to perform logical reasoning and connect relevant concepts, leading to more accurate and interpretable outcomes.

\vspace{-5pt}
\subsection{Ablation Study}
\vspace{-5pt}

In this section, we conduct an ablation study to investigate the contributions of the key components of our Reasoning Scaffolding method, specifically focusing on the hyper-parameter sensitivity analysis on the signal prediction and the impact of signal quality on reasoning performance. For clarity, we present some of the experimental results solely on the 14B model; additional results for other model scales can be found in Appendix~\ref{app:signalQuality} and Appendix~\ref{app:tokenConsume}.

\vspace{-5pt}
\subsubsection{Signal Predictor}
\label{sec:exp:signalPredictor}
\vspace{-5pt}

To enhance the applicability of our method, we train a signal predictor using data derived from golden semantic signals. The training data consists of pairs in the format \text{<current context, next signal>}, constructed as described in Section~\ref{sec:data}. Detailed examples of this training data can be found in Appendix~\ref{app:examples}. We employ supervised fine-tuning for the signal predictor, utilizing the same base model as the proposer, as outlined in Section~\ref{sec:training}. 

As described in Section~\ref{sec:data}, we annotate reasoning steps using a two-stage process that combines keyword matching with LLM-based verification. Empirically, we find that approximately 74\% of all reasoning steps begin with one of our predefined keywords. For these keyword-initiated steps, the labeling accuracy—defined as the proportion of steps for which the keyword-based semantic signal matches the LLM’s semantic labeling—is about 87\%. The remaining 26\% of reasoning steps, which do not start with any predefined keyword, are labeled directly using the LLM oracle (GPT-4.1). Further details on the labeling procedure are provided in Appendix~\ref{app:signalLabeling}.

\begin{table*}[ht]
\centering
\scriptsize
\renewcommand{\arraystretch}{1.1}
\caption{Signal prediction accuracy.}
\begin{tabular}{lcccccc}
\hline
\multicolumn{1}{c}{\textbf{Model Size}} & \textbf{StrategyQA} & \textbf{CommonsenseQA} & \textbf{TruthfulQA} & \textbf{GSM8K} & \textbf{MATH-500} & \textbf{Averaged} \\ \hline
Qwen 0.5B                               & 0.748               & 0.739                  & 0.732               & 0.729          & 0.737             & 0.737             \\
\quad + Adaptive         & 0.783               & 0.777                  & 0.791               & 0.778          & 0.784             & 0.783             \\
Qwen 7B                                 & 0.791               & 0.788                  & 0.802               & 0.801          & 0.796             & 0.796             \\
\quad + Adaptive         & 0.839               & 0.841                  & 0.837               & 0.835          & 0.837             & 0.838             \\
Qwen 14B                                & 0.841               & 0.843                  & 0.841               & 0.829          & 0.836             & 0.838             \\
\quad + Adaptive         & 0.857               & 0.849                  & 0.855               & 0.858          & 0.849             & 0.854             \\ \hline
\end{tabular}
\label{tab:signalPredictor}
\end{table*}

As shown in Table~\ref{tab:signalPredictor}, the signal predictor achieves a next-signal prediction accuracy exceeding 75\%, and this accuracy increases to over 83\% when the base model scale is expanded to 14B. Detailed accuracy rates for each individual signal are reported in Table~\ref{tab:app:signalAcc} in the Appendix. Furthermore, by incorporating the adaptive signal prediction strategy introduced in Section~\ref{sec:inference}, the Pass@1 accuracy for next-signal prediction rises to above 85\%, indicating that reliable and accurate semantic signals can be provided to guide the decoding of proposer models.

\begin{figure*}[ht]
    \centering
    \makebox[\textwidth][c]{\includegraphics[width=0.55\textwidth]{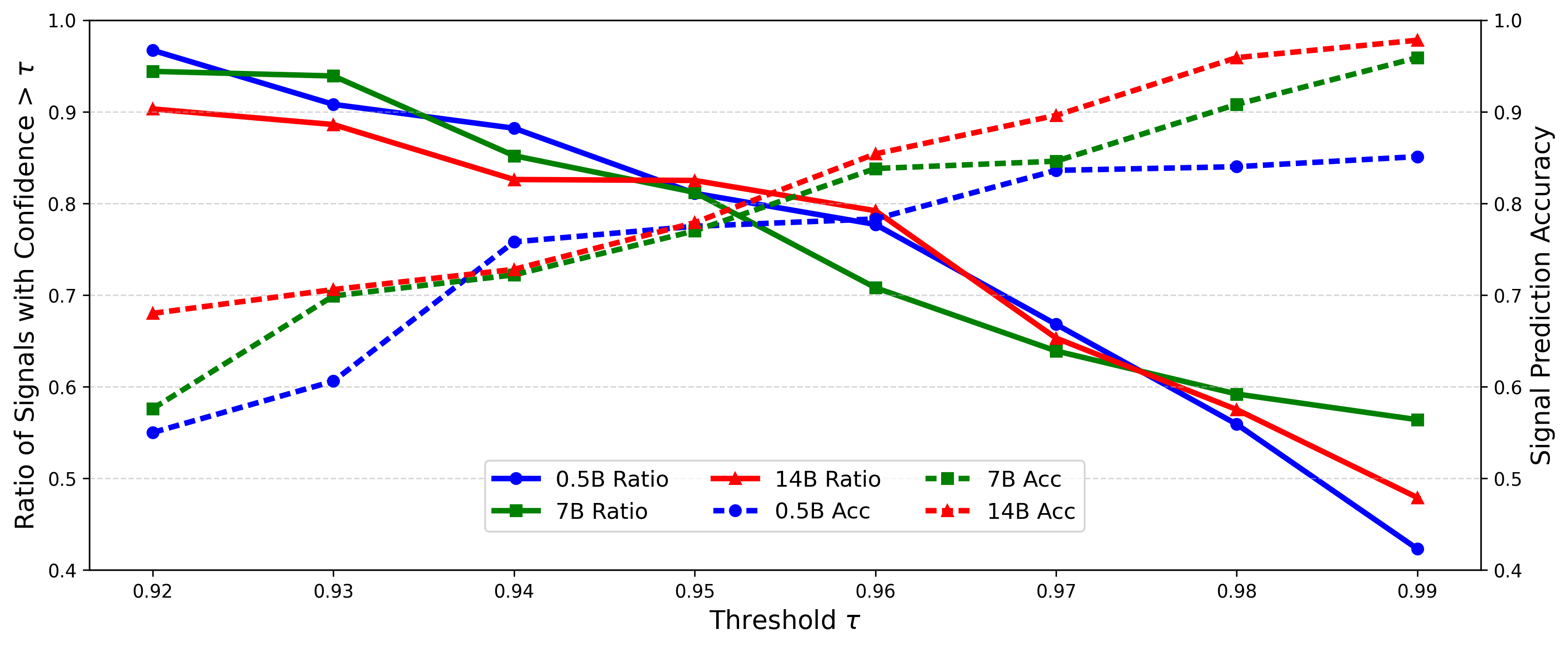}}
    \vspace{-15pt}
    \caption{The effects of threshold $\tau$ on the ratio of predicted signals with log probability $> \tau$ and the signal prediction accuracy.}
    \label{fig:signalTau}
\end{figure*}

Since the adaptive signal prediction strategy is controlled by the threshold parameter $\tau$, we further investigate its impact on prediction accuracy. As illustrated in Figure~\ref{fig:signalTau}, there is a noticeable intersection between the curves representing signal prediction accuracy and the ratio of signals with confidence $> \tau$ for $\tau$ values between 0.95 and 0.96. To strike a balance between retaining a higher proportion of predicted signals and maintaining high prediction accuracy, we set the global value of $\tau$ to 0.96 throughout our experiments.

\subsubsection{Signal Quality and Reasoning Performance}

Table~\ref{tab:signalQuality} illustrates the impact of signal quality on the reasoning performance of the proposer model. Specifically, we conduct three sets of experiments, where the proposer is guided by golden signals, signals generated by the signal predictor, and randomly generated signals. The results clearly show that the quality of signals significantly influences the model's reasoning capabilities.

\vspace{-15pt}
\begin{table*}[ht]
\centering
\scriptsize
\renewcommand{\arraystretch}{1.1}
\setlength{\tabcolsep}{4pt}
\caption{Accuracy analysis under different signal strategies.}
\begin{tabular}{@{}cccccc@{}}
\toprule
\textbf{Benchmarks}    & \textbf{Original} & \textbf{w/ Golden Signals} & \textbf{w/ Signal Predictor} & \textbf{w/ Random Signals} & \textbf{Summaries Only} \\ \midrule
\textbf{StrategyQA}    & 0.755             & 0.858                      & 0.843                        & 0.776                      & 0.855                   \\
\textbf{CommonsenseQA} & 0.785             & 0.887                      & 0.869                        & 0.827                      & 0.869                   \\
\textbf{TruthfulQA}    & 0.750             & 0.917                      & 0.885                        & 0.828                      & 0.897                   \\
\textbf{GSM8K}         & 0.921             & 0.942                      & 0.933                        & 0.929                      & 0.941                   \\
\textbf{MATH-500}      & 0.764             & 0.928                      & 0.918                        & 0.894                      & 0.916                   \\ \bottomrule
\end{tabular}
\label{tab:signalQuality}
\end{table*}
\vspace{-5pt}

When using golden signals, the proposer achieves the highest accuracy across all benchmarks. There is a slight decrease in performance when switching to signals predicted by the signal predictor; however, the accuracy remains substantially higher than the original model. This indicates that the signal predictor is able to reproduce most of the high-quality semantic signal traces.

The strong performance of the `Summaries Only' strategy highlights a key insight: identifying and generating intermediate conclusions (`Conclusion and Summary' steps) is the most critical part of the reasoning scaffold. While the full spectrum of signals provides the best performance, this finding suggests that focusing on conclusion states is a primary driver of the gains.


Interestingly, using random signals still provides a benefit over standard fine-tuning. This suggests that the mere act of structuring the generation into discrete steps—even with a nonsensical scaffold—serves as a useful inductive bias, preventing the model from defaulting to monolithic text imitation. However, using the correct, golden signals provides a significant additional gain, demonstrating that learning the correct reasoning flow is crucial for optimal performance. We also present a failure case where the proposer is guided by shuffled semantic signals, as shown in Appendix~\ref{app:compRandom}.

\subsubsection{Token Consumption Analysis}

Table~\ref{tab:tokenconsume} reports the token consumption across different methods and benchmarks. As expected, token usage increases from CoT-based methods to long-thinking approaches. Our Reasoning Scaffolding method, when guided by all signals, produces reasoning traces with token lengths comparable to those of long-thinking traces distilled from large reasoning models, yet achieves significantly higher reasoning accuracy (see Table~\ref{tab:mainresults}). 

\vspace{-15pt}
\begin{table*}[ht]
\centering
\scriptsize
\renewcommand{\arraystretch}{1.1}
\caption{Token consumption analysis under different strategies.}
\begin{tabular}{lccccc}
\hline
\multicolumn{1}{c}{\textbf{Methods}}          & \textbf{StrategyQA} & \textbf{CommonsenseQA} & \textbf{TruthfulQA} & \textbf{GSM8K} & \textbf{MATH-500} \\
\hline
Original                                      & 224                 & 217                    & 282                 & 304            & 604               \\
CoT SFT                                       & 438                 & 392                    & 509                 & 343            & 715               \\
Thinking Distill                              & 597                 & 613                    & 705                 & 771            & 2,616             \\
Thinking SFT                                  & 1,878               & 1,959                  & 1,877               & 1,996          & 5,921             \\
\hline
\textbf{Ours} + All Signals                   & 1,524               & 1,638                  & 1,550               & 1,659          & 4,755             \\
\quad - Remove Reasoning Steps                & 746                 & 823                    & 773                 & 845            & 2,628             \\
\hline
\end{tabular}
\label{tab:tokenconsume}
\end{table*}
\vspace{-5pt}

We acknowledge that our Reasoning Scaffolding method, in its current form, generates longer reasoning traces compared to standard CoT. This reflects a trade-off where our approach prioritizes maximal reasoning fidelity and logical coherence over token efficiency. The token reduction strategy in Section~\ref{sec:inference} helps mitigate this, the length of reasoning traces can be significantly shortened, while still maintaining high accuracy (refer to the last column of Table~\ref{tab:signalQuality}). future work could explore methods for pruning the scaffold to create more compact yet equally effective reasoning paths.

\section{Limitations and Future Work}
Despite the promising results and novel perspective introduced in this work, several limitations merit discussion for a balanced and transparent account.

First, our approach primarily extracts high-level discourse markers (e.g., Contrast, Addition, Conclusion) rather than formal algorithmic or logical operations. Thus, our `reasoning scaffolding' acts as a tractable and interpretable proxy for reasoning structure, but does not directly encode programmatic or symbolic computation. Future work could address this by incorporating more fine-grained logical or algorithmic signals to bridge the gap between discourse-level and formal reasoning.

Second, our annotation methodology combines heuristic techniques (e.g., segmentation by double newlines, keyword matching) with LLM-based oracle validation (GPT-4.1). While effective, this introduces some dependency on teacher formatting and external models, potentially affecting scalability and reproducibility. Future research could focus on developing more robust, self-supervised signal extraction methods or reducing reliance on costly oracles.

Overall, these limitations do not diminish the core contributions of this work but instead highlight promising directions for future research. Addressing these challenges will further improve the scalability, robustness, and practical utility of reasoning scaffolds in language models.

\section{Conclusion}

We proposed Reasoning Scaffolding, a novel distillation framework that empowers Small Language Models to internalize the structured reasoning patterns of Large Language Models. Unlike conventional rationale-based approaches, our method distills semantic signals rather than surface-level text, effectively addressing key limitations in existing techniques. This leads to student models that are not only more accurate, but also exhibit greater logical consistency and interpretability. Extensive experiments across multiple benchmarks demonstrate the effectiveness of our approach, highlighting its potential as a promising direction for efficient and faithful knowledge transfer in language model distillation.
\newpage




\bibliography{iclr2026_conference}
\bibliographystyle{iclr2026_conference}


\newpage

\section{Appendix}

\subsection{The Use of Large Language Models}
We use Large Language Models (LLMs), including ChatGPT and Gemini, solely for polishing the writing of this paper.

\subsection{Implementation Details}
\label{app:details}

Table~\ref{tab:modelData} summarizes the models and datasets used in our study, including three Qwen2.5-Instruct variants evaluated on five benchmarks with varying train and test set sizes.

\vspace{-15pt}
\begin{table*}[ht]
\centering
\scriptsize
\setlength{\tabcolsep}{15pt}
\caption{Details of models and datasets.}
\begin{tabular}{@{}llllll@{}}
\toprule
\multicolumn{1}{c}{\textbf{Items}} & \multicolumn{5}{c}{\textbf{Values}}                                 \\ \midrule
\textbf{Models}                    & Qwen2.5-0.5B-Ins & Qwen2.5-7B-Ins & Qwen2.5-14B-Ins &       &       \\
\textbf{Benchmarks}                & StrategyQA       & CommonsenseQA  & TruthfulQA      & GSM8K & MATH  \\
\textbf{Train Set Size}            & 1,602            & 9,741          & 657             & 7,473 & 7,500 \\
\textbf{Test Set Size}             & 687              & 1,140          & 160             & 1,319 & 500   \\ \bottomrule
\end{tabular}
\label{tab:modelData}
\end{table*}

\vspace{-15pt}
\begin{table*}[ht]
\centering
\scriptsize
\setlength{\tabcolsep}{14pt}
\caption{Training details.}
\begin{tabular}{@{}lll@{}}
\toprule
\multicolumn{1}{c}{\textbf{Parameter}} & \multicolumn{1}{c}{\textbf{Value}}      & \multicolumn{1}{c}{\textbf{Description}}         \\ \midrule
\textbf{Model Name}                    & Qwen/Qwen2.5-\{0.5B, 7B, 14B\}-Instruct & Base model                                       \\
\textbf{Learning Rate}                 & 1e-4/5e-5                               & Initial learning rate                            \\
\textbf{Batch Size}                    & 1                                       & Per-device batch size                            \\
\textbf{Num Train Epochs}              & 3$\sim$5                                & Number of training epochs                        \\
\textbf{Gradient Accum. Steps}         & 8                                       & Gradient accumulation steps                      \\
\textbf{PEFT / LoRA}                   & True (r=16, $\alpha$=16)                & Adapter fine-tuning (LoRA)                       \\
\textbf{LR Scheduler Type}             & Cosine                                  & Learning Rate Scheduler Type: (Cosine or Linear) \\
\textbf{Completion only Loss}          & True                                    & Compute loss on the completion segment only      \\
\textbf{Warmup Ratio}                  & 0.05                                    & Warmup Ratio                                     \\ \bottomrule
\end{tabular}
\label{tab:trainingParameters}
\end{table*}

Table~\ref{tab:signals} presents all the keywords found in the long-thinking trace along with their corresponding semantic signal names.

\vspace{-15pt}
\begin{table*}[ht]
\centering
\scriptsize
\setlength{\tabcolsep}{3pt}
\caption{All keywords in the long-think trace and their corresponding semantic signal names}
\begin{tabular}{@{}clllllll@{}}
\toprule
\textbf{Semantic Signals}                             & \multicolumn{7}{c}{\textbf{Keywords}}                                                                       \\ \midrule
\multirow{2}{*}{\textbf{Contrast and Concession}}     & but         & however      & on the other hand & otherwise     & nevertheless & nonetheless & in contrast   \\
                                                      & still       & although     & whereas           &               &              &             &               \\ \cmidrule(l){2-8} 
\textbf{Addition and Elaboration}                     & also        & moreover     & additionally      & furthermore   & in addition  &             &               \\ \cmidrule(l){2-8} 
\textbf{Examples and Illustration}                    & for example & for instance &                   &               &              &             &               \\ \cmidrule(l){2-8} 
\multirow{2}{*}{\textbf{Personal Opinion and Recall}} & i think     & i believe    & i guess           & in my opinion & maybe        & it seems    & perhaps       \\
                                                      & i recall    & i remember   &                   &               &              &             &               \\ \cmidrule(l){2-8} 
\multirow{2}{*}{\textbf{Reasoning and Analysis}}      & first       & actually     & in fact           & let me        & anyway       & by the way  & of course     \\
                                                      & i'll/need   & let's see    & wait              & ok            & well         & now         &               \\ \cmidrule(l){2-8} 
\multirow{2}{*}{\textbf{Conclusion and Summary}}      & so          & then         & after all         & obviously     & clearly      & indeed      & meanwhile     \\
                                                      & similarly   & unless       & as a result       & therefore     & thus         & to conclude & in conclusion \\ \bottomrule
\end{tabular}
\label{tab:signals}
\end{table*}

In addition to the aforementioned semantic signals, we introduce another signal, `Response Generation', to indicate the initiation of summary generation for the reasoning trace. From a content perspective, this signal corresponds to the chain-of-thought segment following the `</think>' markers. The `Response Generation' signal is also utilized for early stopping and for directly generating the response summary, as discussed in Section~\ref{sec:inference}.

\begin{table*}[ht]
\centering
\scriptsize
\renewcommand{\arraystretch}{1.1}
\setlength{\tabcolsep}{1pt}
\caption{Keyword covering rate and labeling correctness.}
\begin{tabular}{@{}lccccc@{}}
\toprule
                                       & \textbf{StrategyQA} & \textbf{CommonsenseQA} & \textbf{TruthfulQA} & \textbf{GSM8K} & \textbf{MATH} \\ \midrule
\textbf{Keyword Covering Rate}         & 0.771               & 0.715                  & 0.687               & 0.743          & 0.798         \\
\textbf{Keyword Labeling Correctness} & 0.850               & 0.825                  & 0.922               & 0.874          & 0.903         \\ \bottomrule
\end{tabular}
\label{tab:app:keywordAcc}
\end{table*}
\vspace{-10pt}

\subsection{Experimental Results}

\subsubsection{Signal Labeling Correctness}
\label{app:signalLabeling}

Here we show more details on the data generation accuracy for signal prediction, mainly focusing on the combination of the keyword-based and LLM-based method.

\subsubsection{Detailed Signal Prediction Accuracy}
\label{app:eachSignalAcc}
We also conducted additional experiments to evaluate the prediction accuracy for each individual signal in detail. Table~\ref{tab:app:signalAcc} presents the detailed prediction accuracy for each signal using different base models for training the signal predictor.

\vspace{-10pt}
\begin{table*}[ht]
\centering
\scriptsize
\renewcommand{\arraystretch}{1.1}
\setlength{\tabcolsep}{1pt}
\caption{Signal prediction accuracy of each signal.}
\begin{tabular}{@{}ccccccccc@{}}
\toprule
\textbf{Signal Predictor} & \textbf{\begin{tabular}[c]{@{}c@{}}Reasoning\\ and Analysis\end{tabular}} & \textbf{\begin{tabular}[c]{@{}c@{}}Addition and\\ Elaboration\end{tabular}} & \textbf{\begin{tabular}[c]{@{}c@{}}Examples and\\ Illustration\end{tabular}} & \textbf{\begin{tabular}[c]{@{}c@{}}Personal\\ Opinion and Recall\end{tabular}} & \textbf{\begin{tabular}[c]{@{}c@{}}Contrast and\\ Concession\end{tabular}} & \textbf{\begin{tabular}[c]{@{}c@{}}Conclusion\\ and Summary\end{tabular}} & \textbf{\begin{tabular}[c]{@{}c@{}}Response\\ Generation\end{tabular}} & \textbf{Average} \\ \midrule
Qwen2.5-0.5B-Instruct     & 0.745                                                                     & 0.661                                                                       & 0.851                                                                        & 0.693                                                                          & 0.748                                                                      & 0.822                                                                     & 0.943                                                                  & 0.783            \\
Qwen2.5-7B-Instruct       & 0.815                                                                     & 0.689                                                                       & 0.896                                                                        & 0.767                                                                          & 0.802                                                                      & 0.848                                                                     & 0.979                                                                  & 0.838            \\
Qwen2.5-14B-Instruct      & 0.835                                                                     & 0.721                                                                       & 0.912                                                                        & 0.757                                                                          & 0.816                                                                      & 0.875                                                                     & 0.998                                                                  & 0.854            \\ \bottomrule
\end{tabular}
\label{tab:app:signalAcc}
\end{table*}
\vspace{-5pt}

\subsubsection{Signal Quality and Reasoning Performance (on 0.5B and 7B Model)}
\label{app:signalQuality}

To further validate the impact of signal quality on reasoning performance, we replicate the main experiments on both a smaller 0.5B parameter model and a larger 7B parameter model. Tables~\ref{tab:signalQuality-0.5B} and~\ref{tab:signalQuality-7B} report the accuracy results under five signal strategies across all benchmarks.

\vspace{-10pt}
\begin{table*}[ht]
\centering
\scriptsize
\renewcommand{\arraystretch}{1.1}
\setlength{\tabcolsep}{4pt}
\caption{Accuracy analysis under different signal strategies (0.5B Model).}
\begin{tabular}{cccccc}
\hline
\textbf{Benchmarks}    & \textbf{Original} & \textbf{w/ Golden Signals} & \textbf{w/ Signal Predictor} & \textbf{w/ Random Signals} & \textbf{Summaries Only} \\ \hline
\textbf{StrategyQA}    & 0.543             & 0.659                      & 0.632                        & 0.583                      & 0.64                    \\
\textbf{CommonsenseQA} & 0.475             & 0.601                      & 0.587                        & 0.459                      & 0.601                   \\
\textbf{TruthfulQA}    & 0.267             & 0.861                      & 0.834                        & 0.676                      & 0.855                   \\
\textbf{GSM8K}         & 0.379             & 0.488                      & 0.459                        & 0.39                       & 0.478                   \\
\textbf{MATH-500}      & 0.335             & 0.417                      & 0.393                        & 0.362                      & 0.401                   \\ \hline
\end{tabular}
\label{tab:signalQuality-0.5B}
\end{table*}

Results show that, similar to the main model, the use of golden signals and predicted signals significantly improves accuracy over the original baseline. Notably, even randomly generated signals and summaries-only strategies yield consistent gains, particularly on TruthfulQA and StrategyQA, suggesting the robustness of the Reasoning Scaffolding approach even in smaller models.

\begin{table*}[h!]
\centering
\scriptsize
\renewcommand{\arraystretch}{1.1}
\setlength{\tabcolsep}{4pt}
\caption{Accuracy analysis under different signal strategies (7B Model).}
\begin{tabular}{cccccc}
\hline
\textbf{Benchmarks}    & \textbf{Original} & \textbf{w/ Golden Signals} & \textbf{w/ Signal Predictor} & \textbf{w/ Random Signals} & \textbf{Summaries Only} \\ \hline
\textbf{StrategyQA}    & 0.726             & 0.832                      & 0.797                        & 0.757                      & 0.799                   \\
\textbf{CommonsenseQA} & 0.785             & 0.866                      & 0.841                        & 0.823                      & 0.858                   \\
\textbf{TruthfulQA}    & 0.706             & 0.879                      & 0.859                        & 0.752                      & 0.881                   \\
\textbf{GSM8K}         & 0.875             & 0.899                      & 0.873                        & 0.929                      & 0.882                   \\
\textbf{MATH-500}      & 0.738             & 0.922                      & 0.883                        & 0.866                      & 0.908                   \\ \hline
\end{tabular}
\label{tab:signalQuality-7B}
\end{table*}

The trends closely mirror those observed in the main experiments, with golden signals yielding the highest accuracy across benchmarks. The 7B model demonstrates overall stronger performance, and signal-guided reasoning consistently boosts accuracy, highlighting the scalability and effectiveness of semantic signal guidance across model sizes.

\subsubsection{Token Consumption Analysis (on 0.5B and 7B Models)}
\label{app:tokenConsume}

To assess the efficiency of Reasoning Scaffolding across different model scales, we report token consumption results for both the 0.5B and 7B parameter models under various reasoning strategies. Tables~\ref{tab:tokenconsume-0.5B} and~\ref{tab:tokenconsume-7B} present the average token usage per benchmark for each method.

\begin{table*}[ht]
\centering
\scriptsize
\renewcommand{\arraystretch}{1.1}
\caption{Token consumption analysis under different strategies (0.5B Model).}
\begin{tabular}{lccccc}
\hline
\multicolumn{1}{c}{\textbf{Methods}}          & \textbf{StrategyQA} & \textbf{CommonsenseQA} & \textbf{TruthfulQA} & \textbf{GSM8K} & \textbf{MATH-500} \\ \hline
Original                                      & 180                 & 192                    & 259                 & 271            & 523               \\
CoT SFT                                       & 499                 & 344                    & 449                 & 321            & 679               \\
Thinking SFT                                  & 1797                & 1927                   & 1847                & 1971           & 5652              \\ \hline
\textbf{Ours} + All Signals  & 1447                & 1501                   & 1432                & 1696           & 4673              \\
\quad - Remove Reasoning Steps & 727                 & 749                    & 616                 & 793            & 2235              \\ \hline
\end{tabular}
\label{tab:tokenconsume-0.5B}
\end{table*}

The results highlight that Reasoning Scaffolding (Ours + All Signals) produces longer reasoning traces than standard CoT approaches, but remains more efficient than Thinking SFT. Further token savings are achieved by removing reasoning steps after intermediate conclusions, demonstrating flexibility in controlling output length while maintaining reasoning performance. Compared to the long-thinking method, our Reasoning Scaffolding method can achieve higher performance with a much shorter reasoning trace.

\begin{table*}[ht]
\centering
\scriptsize
\renewcommand{\arraystretch}{1.1}
\caption{Token consumption analysis under different strategies (7B Model).}
\begin{tabular}{lccccc}
\hline
\multicolumn{1}{c}{\textbf{Methods}}          & \textbf{StrategyQA} & \textbf{CommonsenseQA} & \textbf{TruthfulQA} & \textbf{GSM8K} & \textbf{MATH-500} \\ \hline
Original                                      & 216                 & 213                    & 276                 & 320            & 553               \\
CoT SFT                                       & 464                 & 364                    & 491                 & 355            & 704               \\
Thinking Distill                              & 530                 & 572                    & 651                 & 686            & 2422              \\
Thinking SFT                                  & 1840                & 2072                   & 1856                & 2013           & 5841              \\ \hline
\textbf{Ours} + All Signals  & 1579                & 1587                   & 1467                & 1790           & 5103              \\
\quad - Remove Reasoning Steps & 758                 & 819                    & 792                 & 767            & 2422              \\ \hline
\end{tabular}
\label{tab:tokenconsume-7B}
\end{table*}

The trends are consistent with those observed in smaller models: Reasoning Scaffolding delivers detailed reasoning with moderate token growth, and applying the step-removal strategy significantly reduces token usage. This confirms the scalability and efficiency of the approach across model sizes.

\subsubsection{Reasoning Traces Guided by Random Signals}
\label{app:compRandom}
Below, we present a comparison between reasoning traces guided by the original semantic signal order and those guided by a random order. 

The key difference between the two reasoning traces is that the correct answer’s reasoning is grounded in accurate legal facts about the U.S. criminal justice system, specifically that only the prosecutor (not the victim) can drop criminal charges once filed, and each step checks assumptions against legal reality. In contrast, the first reasoning trace relies on personal belief and incorrect assumptions—mistakenly thinking the victim has the authority to drop charges—without verifying this against actual legal procedure. The first reasoning fails because it builds its logic on a fundamental factual error about who controls criminal prosecution, never fact-checking this critical point.

\begin{tcolorbox}[colback=black!3!white, colframe=black!30!white, center title,left=5pt,right=5pt,breakable]
    \begin{tcolorbox}[colback=black!3!white, colframe=black!30!white, boxrule=0.5mm]
        \begin{center}
            Question
        \end{center}
        \scriptsize
        In the U.S. can the victim of a domestic violence case drop the charges? Options: A. Yes, the victim can drop the charges B.  Yes, if the victim is the accuser, they are free to drop the charges C. No, the victim cannot drop the charges once a prosecutor is involved
    \end{tcolorbox}
    \begin{tcolorbox}[colback=black!3!white, colframe=black!30!white, boxrule=0.5mm,left=1pt, right=1pt,top=1pt,bottom=1pt]
        \begin{center}
            Original Semantic Signals
        \end{center}
        \scriptsize
Reasoning and Analysis, Reasoning and Analysis, Contrast and Concession, Personal Opinion and Recall, Contrast and Concession, Contrast and Concession, Personal Opinion and Recall, Contrast and Concession, Reasoning and Analysis, Examples and Illustration, Conclusion and Summary, Contrast and Concession, Personal Opinion and Recall, Contrast and Concession, Reasoning and Analysis, Personal Opinion and Recall, Conclusion and Summary, Conclusion and Summary, Conclusion and Summary, Conclusion and Summary, Contrast and Concession
    \end{tcolorbox}
    \begin{tcolorbox}[colback=black!3!white, colframe=black!30!white, boxrule=0.5mm,left=1pt, right=1pt,top=1pt,bottom=1pt]
        \begin{center}
            Random Semantic Signals
        \end{center}
        \scriptsize
Personal Opinion and Recall, Examples and Illustration, Conclusion and Summary, Contrast and Concession, Contrast and Concession, Personal Opinion and Recall, Conclusion and Summary, Reasoning and Analysis, Personal Opinion and Recall, Contrast and Concession, Conclusion and Summary, Contrast and Concession, Conclusion and Summary, Contrast and Concession, Conclusion and Summary, Conclusion and Summary, Reasoning and Analysis, Contrast and Concession, Reasoning and Analysis, Contrast and Concession, Reasoning and Analysis
    \end{tcolorbox}
    \begin{minipage}{0.38\textwidth}
        \begin{tcolorbox}[colback=black!3!white, colframe=black!30!white,left=1pt,right=1pt]
        \begin{center}
            Reasoning Trace Guided by Random Signals
        \end{center}
        \scriptsize

1. Personal Opinion and Recall: \textcolor{red}{I think I remember that in the US, the victims of domestic violence have the right to drop charges.}

2. Examples and Illustration: \textcolor{red}{For example, in many jurisdictions, the prosecution cannot overcharge or force the case to trial if the victim doesn't want it.}

3. Conclusion and Summary: \textcolor{red}{So, option A seems correct.}

4. Contrast and Concession: But let's look at option B: Yes, if the victims is the accused, they are free to drop the charges That seems redundant because if the same person is both the victims and the accused, it might be confusing, but essentially, they can still drop charges.

5. Contrast and Concession: But option C says: "No, the victims cannot drop the charges once a prosecuted is involved" This might be tricky

6. Personal Opinion and Recall: \textcolor{red}{I recall that when a case is prosecuted, the prosecution can apply for a warrant or move the case along, but the final decision to drop charges is often with the victims, especially in domestic violence cases where there might be a lot of personal details.}

7. Conclusion and Summary: \textcolor{red}{I think A is correct.}

8. Reasoning and Analysis: Wait, let me double-check.

9. Personal Opinion and Recall: \textcolor{red}{I think in some states, there might be specific laws, but generally, the principle is that the victims have a say in the case} Option B says if the victims is the accused, which might mean the same person is both, but it's poorly worded It should probably be if the victim is the accused, but even then, it's not clear.

10. Contrast and Concession: But in context, it's probably meant to be the same as A.

11. Conclusion and Summary: \textcolor{red}{So, I think A is the best answer.}

12. Contrast and Concession: But let's see why C might be tempting. Sometimes, if the charge is serious, like if a warrant is out, the prosecution might not allow it, but for domestic violence, it's common to have the option to drop.

13. Conclusion and Summary: \textcolor{red}{Therefore, I think A is correct.}

14. Contrast and Concession: But just to be thorough, I remember that in some cases, like if the crime is perjury or something, but for domestic violence, it's fine.

15. Conclusion and Summary: \textcolor{red}{So, my final answer should be A.}

16. Conclusion and Summary: \textcolor{red}{So, choose A.}

17. Reasoning and Analysis: OK, I think that's it.

18. Contrast and Concession: But let's see the options again.

19. Reasoning and Analysis: Actually, A Yes, the victims can drop the charges B Yes, if the victims is the accused, they are free to drop the charges – this might imply that in some cases, the accused is also the victims, but it's awkward phrasing C No, the victims cannot drop the charges once a prosecuted is involved – this suggests that once involved, they can't drop, which is false.

20. Contrast and Concession: \textcolor{red}{But in reality, they can.}

21. Reasoning and Analysis: \textcolor{red}{OK, A is correct.}

        \end{tcolorbox}
    \end{minipage}
    \hfill
    \begin{minipage}{0.62\textwidth}
        \begin{tcolorbox}[colback=black!3!white, colframe=black!30!white,left=1pt,right=1pt]
        \begin{center}
        Reasoning Trace Guided by Godel Signals
        \end{center}
        \scriptsize
        
1. Reasoning and Analysis: First, the question is: "In the U.S. can the victim of a domestic violence case drop the charges?" The options are:A. Yes, the victim can drop the charges B. Yes, if the victim is the accuser, they are free to drop the charge C. No, the victim cannot drop the charges once a prosecutor is involved.

2. Reasoning and Analysis: I need to recall how criminal charges work in the U.S. especially for domestic violence. In criminal cases, it's not the victim who files charges; it's the state or the government. The victim might report the crime or be a witness, but the prosecution is handled by a prosecutor, like a district attorney. For domestic violence cases, victims often want to drop charges because of fear, reconciliation, or other reasons.

3. Contrast and Concession: But can they actually drop the charges themselves?

4. Personal Opinion and Recall: \textcolor{blue}{I remember that once charges are filed, the victim doesn't have the authority to drop them. The prosecutor decides whether to proceed with the case or not.} \textcolor{blue}{This is because domestic violence is considered a crime against the state, not just the individual victim.} Option C says: "No, the victim cannot drop the charges once a prosecutor is involved." That sounds familiar.

5. Contrast and Concession: But let me check options A and B. Option A says "Yes, the victim can drop the charges," which seems too broad and probably incorrect. Option B says "Yes, if the victim is the accuser, they are free to drop the charges."

6. Contrast and Concession: But in criminal cases, the accuser is typically the state, not the victim. The victim might be the complainant, but not the one prosecuting. In some jurisdictions, victims can file for restraining orders or civil actions, but criminal charges are separate.

7. Personal Opinion and Recall: \textcolor{blue}{I think there's a common misconception that victims can drop charges, but in reality, they cannot. Prosecutors can consider the victim's wishes, but they are not bound by them. Prosecutors might proceed even if the victim wants to drop charges, especially in domestic violence cases, to prevent coercion or further harm.}

8. Contrast and Concession: But is that universal? I think it varies, but generally, victims cannot unilaterally drop charges.

9. Reasoning and Analysis: Let me think about the legal process. When a crime is reported, the police investigate, and if there's enough evidence, the prosecutor files charges. The victim becomes a witness. \textcolor{blue}{The victim doesn't have the power to dismiss the case; only the prosecutor or the court can do that.}

10. Examples and Illustration: For example, if a victim refuses to testify, the prosecution might have a harder time, but they can still proceed with other evidence or subpoena the victim.

11. Conclusion and Summary: \textcolor{blue}{So, option C seems correct: "No, the victim cannot drop the charges once a prosecutor is involved."}

12. Contrast and Concession: But what about before charges are filed? If the victim reports domestic violence but later decides not to pursue it, can they prevent charges from being filed? In some cases, if no charges have been filed yet, the victim might not cooperate with the investigation, which could lead to no charges. But once charges are filed, it's out of their hands. The question says "drop the charges," implying that charges have already been filed. Option B mentions "if the victim is the accuser." In criminal law, the accuser is the state, not the individual. So that might be misleading. In some contexts, like private prosecutions, but I think in the U.S. criminal cases are almost always public, with state prosecutors.

13. Personal Opinion and Recall: I think C is the correct answer.

14. Contrast and Concession: But let me double-check. Is there any scenario where a victim can drop charges? For instance, in some misdemeanor cases or with consent, but for domestic violence, it's often treated more seriously.

15. Reasoning and Analysis: Actually, upon second thought, in some jurisdictions, for certain offenses, victims might have more say, but generally for felonies and serious crimes, no. Domestic violence can be a misdemeanor or felony, but still, the principle holds.

16. Personal Opinion and Recall: \textcolor{blue}{I remember a key point: the victim is not a party to the criminal case; the case is The State v. Defendant, not Victim v. Defendant. So the victim doesn't control the prosecution.}

17. Conclusion and Summary: Therefore, the victim cannot drop charges.

18. Conclusion and Summary: So, options A and B are incorrect. Option B says "if the victim is the accuser," but they aren't the accuser in the legal sense; the prosecutor is.

19. Conclusion and Summary: So, I think C is correct. Final answer should be C. The question is multiple choice, and I need to box it.

20. Conclusion and Summary: So, \textcolor{blue}{\textbackslash boxed{C}}

21. Contrast and Concession: But let me make sure. Is there any trick here? The question says "in the U.S." and laws can vary by state, but generally, this is true across states. Yes, I think it's standard.

        \end{tcolorbox}
    \end{minipage}
\end{tcolorbox}

\subsection{Prompt Design}
\subsubsection{Prompts for Benchmarking}
\label{App:benchmarkPrompt}
Below is a series of prompts used to benchmark the base models for direct answering, zero-shot prompting, and few-shot prompting.

\begin{tcolorbox}[colback=black!3!white,colframe=black!30!white,breakable,title=Prompts for Directly Answering Prompting Strategy for Benchmarking, center title]
\scriptsize

\text{Answer the following single-choice question directly and putting your choice in \#\#\#Answer:}\textbackslash boxed\{\} without any explanations.

This is the question:

\{question\}
\newline

Your choice:

\end{tcolorbox}

\begin{tcolorbox}[colback=black!3!white,colframe=black!30!white,breakable,title=Prompts for Zero-shot Prompting Strategy for Benchmarking,center title]
\scriptsize

\text{Answer the following single-choice question step-by-step and putting your choice in \#\#\#Answer:}\textbackslash boxed\{\}.

This is the question:

\{question\}
\newline

Your solution:

\end{tcolorbox}

\begin{tcolorbox}[colback=black!3!white,colframe=black!30!white,breakable,title=Prompts for Few-shot Prompting Strategy for Benchmarking,center title]
\scriptsize

\text{Answer the following single-choice question step-by-step and putting your choice in \#\#\#Answer:}\textbackslash boxed\{\}.

This is the question:

What is the most common cause of the Bermuda Triangle disappearances? Options: \textbackslash nA. The Bermuda Triangle disappearances are caused by a mysterious alien force.\textbackslash nB. The Bermuda Triangle disappearances are the result of a curse placed on the region.\textbackslash nC. There is no single cause, but factors like magnetic anomalies, human error, and severe weather have contributed to the disappearances.\textbackslash nD. The Bermuda Triangle disappearances are caused by a large underwater vortex.
\newline

Your solution:

Let me answer this question about the Bermuda Triangle disappearances step by step:\textbackslash n\textbackslash n1) First, I need to consider what the Bermuda Triangle is. It's a region in the western part of the North Atlantic Ocean where ships and aircraft have allegedly disappeared under mysterious circumstances.\textbackslash n\textbackslash n2) Let me evaluate each option:\textbackslash n\textbackslash n   - Option A suggests alien forces are responsible. This is a popular theme in science fiction but lacks scientific evidence.\textbackslash n\textbackslash n   - Option B suggests a curse is responsible. This is a supernatural explanation without scientific basis.\textbackslash n\textbackslash n   - Option C suggests multiple natural factors including weather, human error, and magnetic anomalies. This aligns with scientific investigations.\textbackslash n\textbackslash n   - Option D suggests an underwater vortex. While ocean currents exist, there's no evidence of a specific vortex causing disappearances.\textbackslash n\textbackslash n3) Scientific investigations of incidents in the Bermuda Triangle have found:\textbackslash n  - Many disappearances can be explained by severe weather like hurricanes and waterspouts\textbackslash n   - Human error in navigation\textbackslash n   - Equipment failure\textbackslash n   - Some compass variations due to natural magnetic properties of the region\textbackslash n   - Many incidents occurred during conditions of poor visibility\textbackslash n\textbackslash n4) Additionally, the U.S. Coast Guard and other scientific organizations have stated that the number of disappearances in the Bermuda Triangle is not statistically higher than other busy shipping areas when accounting for the high volume of traffic.\textbackslash n\textbackslash n5) The most logical explanation supported by evidence is that multiple natural factors, rather than a single supernatural cause, explain the various incidents.\textbackslash n\textbackslash nTherefore, the answer is: C.\textbackslash n\textbackslash n\#\#\#Answer: \textbackslash boxed\{C\}
\newline

\text{Answer the following single-choice question step-by-step and putting your choice in \#\#\#Answer:}\textbackslash boxed\{\}.

This is the question:

Does the popular saying 'You can't teach an old dog new tricks' have any basis in science? Options: \textbackslash nA. The saying is based on the fact that older dogs' brains shrink, making learning impossible.\textbackslash nB. The saying is true, older dogs can only learn new tricks with intense training.\textbackslash nC. The saying is a misconception, as research shows that older dogs can still learn new tricks, though it may take more time and patience.\textbackslash nD. The saying is true, older dogs are incapable of learning new tricks.
\newline

Your solution:

Let me approach this question about whether "You can't teach an old dog new tricks" has any basis in science.\textbackslash n\textbackslash nStep 1: Analyze what science tells us about aging and learning in dogs.\textbackslash nResearch in canine cognition has shown that while aging does affect cognitive function in dogs (as in humans), older dogs maintain the ability to learn. Their brains don't simply "shrink" and become incapable of forming new neural connections. Cognitive decline can occur, but it doesn't make learning "impossible."\textbackslash n\textbackslash nStep 2: Consider how age affects learning in dogs.\textbackslash nStudies have found that older dogs may:\textbackslash n- Learn more slowly than younger dogs\textbackslash n- Require more repetitions\textbackslash n- Need more patience during training\textbackslash n- Benefit from shorter training sessions\textbackslash n- Experience some decline in certain cognitive functions\textbackslash n\textbackslash nHowever, these changes don't prevent learning altogether - they just change the optimal approach to teaching.\textbackslash n\textbackslash nStep 3: Evaluate the scientific evidence for each option.\textbackslash nA: Incorrect - While some brain atrophy can occur with age, it doesn't make learning "impossible."\textbackslash nB: Incorrect - Older dogs don't require "intense" training; they require appropriate, patient training.\textbackslash nC: Aligns with scientific evidence - Research does show older dogs can learn, though potentially with modified approaches.\textbackslash nD: Incorrect - This contradicts scientific evidence on canine cognition.\textbackslash n\textbackslash nStep 4: Draw a conclusion.\textbackslash nThe saying "You can't teach an old dog new tricks" is indeed a misconception. Science shows that while aging affects learning processes, older dogs maintain the ability to learn throughout their lives. The key difference is that it may take more time, patience, and appropriate teaching methods.\textbackslash n\textbackslash nTherefore, the answer is: C\textbackslash n\textbackslash n\#\#\#Answer: \textbackslash boxed\{C\}
\newline

\text{Answer the following single-choice question step-by-step and putting your choice in \#\#\#Answer:}\textbackslash boxed\{\}.

This is the question:

\{question\}
\newline

Your solution:

\end{tcolorbox}

\subsubsection{Prompts for Generating Long-Thinking Data}
\label{App:promptDG}
We directly prompt the DeepSeek-R1 model to obtain the long-thinking trace as the initial dataset.

\begin{tcolorbox}[colback=black!3!white,colframe=black!30!white,breakable,title=Prompts for Generating Initial Data,center title]
\scriptsize

Answer the following math question step by step and put your final choice in \textbackslash n \#\#\#Answer: \textbackslash\textbackslash boxed\{\}
\newline

This is the question: 
\{question\}
\newline

Your solution:

\end{tcolorbox}

\subsubsection{Prompts for Labeling Reasoning Traces by Semantic Meaning}
\label{App:promptSignal}
To enhance the correctness and reliability of semantic signals as labels for reasoning steps, we also employ LLMs as judges to verify whether the logic signals are assigned appropriately to each reasoning step.

\begin{tcolorbox}[colback=black!3!white,colframe=black!30!white,breakable,title=Prompts for Semantic Signal Checking,center title]
\scriptsize

You are a judge who evaluates the semantic logic transition between a new sentence/paragraph (the "text") and its preceding context.
Select the transition type strictly from the following list:
\newline

\text{1. Contrast and Concession: The new sentence or paragraph presents a contrast or concession in relation to the previous context.}
\text{2. Addition and Elaboration: The new sentence or paragraph supplements or expands upon the previous context.}
\text{3. Examples and Illustration: The new sentence or paragraph provides examples or explanations related to the previous context.}
\text{4. Personal Opinion and Recall: The new sentence or paragraph primarily expresses a personal viewpoint or belief.}

5. Reasoning and Analysis: The new sentence or paragraph serves as a starting point for reasoning or doing analysis based on the current information.

\text{6. Conclusion and Summary: The new sentence or paragraph concludes from the previous context and present an intermediate result.}
\text{\quad outcome.}
\newline

\#\# Instructions:
\text{Given the previous context and the new sentence, analyze the semantic logic transition from the context to the text.}
\text{Your answer should contain only the transition type, formatted as:
\textbackslash boxed\{logic transition name\}}
\newline

\#\# Context:

\text{\{context\}}
\newline

\#\# Text to be judged:

\{text\}
\newline

\#\# Your Response:

\end{tcolorbox}

\subsubsection{Prompts for Signal Prediction}
\label{App:promptSignalPred}
Here is the prompt for training a signal predictor, mainly including the semantic signals and the task description for signal prediction.

\begin{tcolorbox}[colback=black!3!white,colframe=black!30!white,breakable,title=Prompts for Signal Predictor,center title,breakable]
\scriptsize

Given the following context from a reasoning or problem-solving process, predict the most likely semantic signal (from a predefined set of signal groups) to which the next reasoning keyword or phrase will belong.
\newline

The possible semantic signals are:
\newline

\text{1. Contrast and Concession}

\text{2. Addition and Elaboration}

\text{3. Examples and Illustration}

\text{4. Personal Opinion and Recall}

\text{5. Reasoning and Analysis}

\text{6. Conclusion and Summary}

\text{7. Response Generation}
\newline

Please answer by putting the semantic signal name in \textbackslash\textbackslash boxed\{semantic signal name\}.
\newline

This is the reasoning context:

\{context\}
\newline

The semantic signal is:

\end{tcolorbox}

\subsection{Training Examples}
\label{app:examples}

Below, we provide examples used for training and testing the Proposer and the Signal Predictor. Specifically, for training the Proposer, the input is the question, and we update the model’s parameters by computing both the token loss—between the gold reasoning step and the predicted step—and the signal loss—between the gold signal and the predicted signal—using the two branches of the modified model architecture.

\begin{tcolorbox}[colback=black!3!white, colframe=black!30!white, title=Example-1, center title,left=5pt,right=5pt,breakable]
    \begin{tcolorbox}[colback=black!3!white, colframe=black!30!white, boxrule=0.5mm]
        \begin{center}
            Question
        \end{center}
        \scriptsize
        What should you do if you have a nosebleed? Options: A.  You should lie down B. You should lean forward and breathe through your mouth C. You should tilt your head back
    \end{tcolorbox}
    \begin{minipage}{0.69\textwidth}
        \begin{tcolorbox}[colback=black!3!white, colframe=black!30!white,left=5pt,right=5pt]
        \begin{center}
            Step-by-Step Reasoning Trace
        \end{center}
        \scriptsize
1. First, the question is: "What should you do if you have a nosebleed?" with three options: A. You should lie down B. You lean forward and breathe through your mouth C. You should tilt your head back.

2. I need to recall the correct first aid for nosebleeds. I've heard different advice, but I think the standard recommendation has changed over time.

3. Actually, I remember that tilting your head back used to be a common suggestion, but it's not recommended anymore. Why? Because if you tilt your head back, the blood can run down your throat, which might cause choking or swallowing blood, and that can lead to nausea or vomiting. 

4. So, option C: "You should tilt your head back" – that might be incorrect.

5. Now, option A: "You should lie down." Is lying down good for a nosebleed? I think lying down could also cause blood to flow down the throat, similar to tilting the head back. It might increase pressure in the veins of the nose, making bleeding worse. Option B: This sounds familiar. Leaning forward helps the blood drain out through the nose instead of down the throat. Breathing through the mouth prevents you from breathing in blood or causing more irritation.

6. I think the American Red Cross or Mayo Clinic recommends sitting upright and leaning forward slightly. Pinching the soft part of the nose is also key, but that's not mentioned in the options. The question is about what to do, and option B includes leaning forward and breathing through the mouth.

7. Let me confirm: Yes, standard advice is to sit down, lean forward, pinch the nostrils together for about 10 minutes, and breathe through your mouth. 

8. So, option B seems correct.

9. But let's double-check option A and C. Option A: Lie down – no, that's bad because it can cause blood to go into the stomach or lungs. Option C: Tilt head back – same issue, blood goes down the throat. Option B: Lean forward – good, blood comes out, and you can spit it out if needed.

10. Also, breathing through the mouth makes sense to avoid nasal passages.

11. I think B is the correct answer.

12. But is there any scenario where you might lie down or tilt back? I don't think so for typical first aid. Medical professionals might have other methods, but for general public, leaning forward is key. The question says "you should do," implying general advice. 

13. So, I'll go with B 

14. So, final answer should be B.
        \end{tcolorbox}
    \end{minipage}
    \hfill
    \begin{minipage}{0.30\textwidth}
        \begin{tcolorbox}[colback=black!3!white, colframe=black!30!white,left=5pt,right=5pt]
        \begin{center}
        Semantic Signals
        \end{center}
        \scriptsize

1. Reasoning and Analysis,
\newline
\newline

2. Reasoning and Analysis,
\newline

3. Reasoning and Analysis,
\newline
\newline
\newline

4. Conclusion and Summary, 

5. Reasoning and Analysis,
\newline
\newline
\newline
\newline
\newline

6. Personal Opinion and Recall,
\newline
\newline
\newline

7. Reasoning and Analysis, 
\newline

8. Conclusion and Summary,

9. Contrast and Concession,
\newline
\newline
\newline

10. Addition and Elaboration,

11. Personal Opinion and Recall,

12. Contrast and Concession,
\newline
\newline
\newline

13. Conclusion and Summary, 

14. Conclusion and Summary,

        \end{tcolorbox}
    \end{minipage}
\end{tcolorbox}

\newpage

\begin{tcolorbox}[colback=black!3!white, colframe=black!30!white, title=Example-2, center title,left=5pt,right=5pt,breakable]
    \begin{tcolorbox}[colback=black!3!white, colframe=black!30!white, boxrule=0.5mm]
        \begin{center}
            Question
        \end{center}
        \scriptsize
What does it always mean when a cat purrs? Options: A.  A cat's purr always means that the cat is hungry B.  A cat's purr always means that the cat is stressed C. A cat's purr can signify many emotions, including both happiness and anxiety D. A cat's purr always means that the cat is happy
    \end{tcolorbox}
    \begin{minipage}{0.69\textwidth}
        \begin{tcolorbox}[colback=black!3!white, colframe=black!30!white,left=5pt,right=5pt]
        \begin{center}
            Step-by-Step Reasoning Trace
        \end{center}
        \scriptsize
1. First, the question is: "What does it always mean when a cat purrs?" The options are: A. A cat's purr always means that the cat is hungry B. A cat's purr always means that the cat is stressed C. A cat's purr can signify many emotions, including both happiness and anxiety D. A cat's purr always means that the cat is happy

2. I need to choose the correct one and box my answer at the end. Key point: The word "always" is crucial here. It says "always mean," so I need to find which statement is universally true for all purring instances.

3. I recall that cats purr for various reasons. From what I know, purring isn't just about one emotion. Cats purr when they're content, like when they're being petted or resting comfortably. But they also purr when they're in pain, stressed, or even when giving birth. I think purring can be a self-soothing mechanism. 

4. For example, a cat might purr when it's happy and relaxed, not necessarily hungry. Or it might purr when injured, which isn't happiness or hunger.

5. I remember reading that kittens purr when nursing, which might indicate contentment, but adult cats purr in various situations.

6. Also, some studies suggest that purring can promote healing or reduce stress, so it's not just about positive emotions.

7. So, option C seems correct because it acknowledges that purring can mean different things, not just one.

8. But the question says "always mean," and C says "can signify," which might not directly answer "always." Let me read the options again. Option C: "A cat's purr can signify many emotions..." – this implies that it doesn't always mean one thing; it can vary.

9. Whereas A, B, D claim that it always means one specific thing.

10. But based on my knowledge, purring doesn't always mean hunger, stress, or happiness. It depends on the context.

11. Therefore, A, B, and D are incorrect because they are absolute statements that aren't true. C is the only one that correctly states that purring can signify multiple emotions. The question is phrased as "what does it always mean," but C doesn't say what it always means; it says what it can signify. That might be a trick. Let me see.

12. Perhaps C is the best answer because it refutes the idea that it always means one thing. Option C is not saying what it always means; it's saying that it can mean different things. 

13. But the question asks for what it always means, implying a consistent meaning. In reality, there is no single thing that purring always means. It varies. 

14. So, none of A, B, D are correct, but C is the only one that captures the variability. Option C starts with "A cat's purr can signify..." which is not claiming an always meaning, but it's the correct understanding. 

15. Perhaps the question is designed to highlight that purring doesn't have a single always meaning, and C is the accurate choice. 

16. Let me check online or recall reliable sources. 

17. I think scientifically, purring is associated with both positive and negative situations. 

18. For instance, cats purr when they're content but also when they're frightened or in pain. 

19. So, it's not always hunger, not always stress, not always happy. 

20. Therefore, C is the correct answer.  
        \end{tcolorbox}
    \end{minipage}
    \hfill
    \begin{minipage}{0.30\textwidth}
        \begin{tcolorbox}[colback=black!3!white, colframe=black!30!white,left=5pt,right=5pt]
        \begin{center}
        Semantic Signals
        \end{center}
        \scriptsize

1. Reasoning and Analysis
\newline
\newline
\newline

2. Reasoning and Analysis
\newline
\newline

3. Personal Opinion and Recall, 
\newline
\newline
\newline

4. Examples and Illustration,
\newline

5. Personal Opinion and Recall, 
\newline

6. Addition and Elaboration, 
\newline

7. Conclusion and Summary, 
\newline

8. Contrast and Concession, 
\newline
\newline
\newline

9. Contrast and Concession, 

10. Contrast and Concession, 
\newline

11. Conclusion and Summary,
\newline
\newline
\newline

12. Personal Opinion and Recall,
\newline
\newline

13. Contrast and Concession, 
\newline

14. Conclusion and Summary, 
\newline
\newline

15. Personal Opinion and Recall, 
\newline

16. Reasoning and Analysis, 

17. Personal Opinion and Recall, 

18. Examples and Illustration,
\newline

19. Conclusion and Summary, 

20. Conclusion and Summary, 
        \end{tcolorbox}
    \end{minipage}
\end{tcolorbox}

\newpage

\begin{tcolorbox}[colback=black!3!white, colframe=black!30!white, title=Example-3, center title,left=5pt,right=5pt,breakable]
    \begin{tcolorbox}[colback=black!3!white, colframe=black!30!white, boxrule=0.5mm]
        \begin{center}
            Question
        \end{center}
        \scriptsize
Jan has three times the number of pets as Marcia. Marcia has two more pets than Cindy. If Cindy has four pets, how many total pets do the three have?
    \end{tcolorbox}
    \begin{minipage}{0.69\textwidth}
        \begin{tcolorbox}[colback=black!3!white, colframe=black!30!white,left=5pt,right=5pt]
        \begin{center}
            Step-by-Step Reasoning Trace
        \end{center}
        \scriptsize
1. First, the question is: Jan has three times the number of pets as Marcia. Marcia has two more pets than Cindy. Cindy has four pets. I need to find the total number of pets for all three. Cindy has four pets. That's given directly: "Cindy has four pets."

2. Now, Marcia has two more pets than Cindy. So, if Cindy has 4, then Marcia has 4 + 2 = 6 pets. 

3. Let me write that down: - Cindy: 4 pets - Marcia: Cindy's pets + 2 = 4 + 2 = 6 pets Next, Jan has three times the number of pets as Marcia. Marcia has 6 pets, so Jan has 3 times that. 

4. So, Jan: 3 * Marcia's pets = 3 * 6 = 18 pets. 

5. Now, to find the total, I add up all their pets: Cindy: 4 Marcia: 6 Jan: 18 Total: 4 + 6 + 18 

6. Let me calculate that: 4 + 6 is 10, and 10 + 18 is 28. 

7. So, the total should be 28 pets. I should double-check to make sure I didn't miss anything. The relationships: - Jan = 3 * Marcia - Marcia = Cindy + 2 - Cindy = 4. So, from Cindy, Marcia = 4 + 2 = 6 

8. Then Jan = 3 * 6 = 18 Total = 4 + 6 + 18 = 28 I can set up equations to verify. Let C be Cindy's pets. C = 4, M = C + 2, J = 3 * M 

9. Then total T = J + M + C, Substitute: M = C + 2 = 4 + 2 = 6, J = 3 * M = 3 * 6 = 18, T = 18 + 6 + 4 = 28. Or directly: T = J + M + C = 3M + M + C = 4M + C 

10. But M = C + 2, so T = 4(C + 2) + C = 4C + 8 + C = 5C + 8, C = 4, so T = 5*4 + 8 = 20 + 8 = 28. Same thing. 

11. So, it seems consistent. The question is to put the answer in a boxed format. 

12. Also, in my thinking, I should include keywords like "Actually," "Wait," etc. but I think I did it step by step without needing to backtrack, but I can add some for the reflection. 

13. For example, after finding Marcia has 6, I could say "OK, so Marcia has 6 pets." But it's fine.

14. I think the process is clear.

15. So, the total number of pets is 28.
        \end{tcolorbox}
    \end{minipage}
    \hfill
    \begin{minipage}{0.30\textwidth}
        \begin{tcolorbox}[colback=black!3!white, colframe=black!30!white,left=5pt,right=5pt]
        \begin{center}
        Semantic Signals
        \end{center}
        \scriptsize
1. Reasoning and Analysis, 
\newline
\newline

2. Reasoning and Analysis, 
\newline

3. Reasoning and Analysis, 
\newline
\newline

4. Conclusion and Summary, 

5. Reasoning and Analysis, 
\newline

6. Reasoning and Analysis, 

7. Conclusion and Summary,
\newline
\newline

8. Conclusion and Summary, 
\newline

9. Conclusion and Summary, 
\newline

10. Contrast and Concession,
\newline

11. Conclusion and Summary, 

12. Addition and Elaboration,
\newline
\newline

13. Examples and Illustration, 
\newline

14. Personal Opinion and Recall, 

15. Conclusion and Summary
        \end{tcolorbox}
    \end{minipage}
\end{tcolorbox}

\subsection{Broader Impact}
The development of Reasoning Scaffolding offers significant potential to reshape both the technical landscape of language modeling and the broader societal deployment of AI systems. By enabling small language models (SLMs) to internalize algorithmic reasoning structures rather than imitate surface-level text, our framework advances knowledge distillation toward greater logical robustness, interpretability, and resource efficiency. This approach directly addresses key challenges in AI safety and reliability, yielding SLMs that are less susceptible to spurious correlations and brittle failure modes, and facilitating trustworthy applications in domains such as education, healthcare, legal reasoning, and scientific discovery. By enhancing the reasoning abilities of compact models, Reasoning Scaffolding democratizes access to advanced AI, supporting robust performance on edge devices and in resource-limited environments. Furthermore, by shifting distillation from rote imitation to structured reasoning, our work paves the way for future research into more transparent and faithful knowledge transfer, including finer-grained reasoning signals, symbolic integration, and self-supervised techniques to further improve robustness and reduce reliance on proprietary LLMs.

\end{document}